\documentclass{iosart2c}

\usepackage{amssymb}
\usepackage{graphicx,wrapfig,lipsum,mathtools}
\usepackage{graphicx,amsmath}
\usepackage{graphics,color,csquotes}
\usepackage{tabularx}
\usepackage{booktabs,colortbl}
\usepackage{multirow,csquotes}
\usepackage[table,x11names]{xcolor}
\usepackage{url}
\begin{document}
\begin{frontmatter}                           % The preamble begins here.

\title{Difficulty-level Modeling of Ontology-based Factual Questions}
\runningtitle{Instructions for the preparation of a camera-ready paper in \LaTeX}
%\subtitle{Subtitle}

\review{Name Surname, University, Country}{Name Surname, University, Country}{Name Surname, University, Country}

\author{\fnms{Vinu} \snm{E.V}\thanks{Corresponding author. E-mail: vinuev@cse.iitm.ac.in, mvsquare1729@gmail.com}}
and
\author{\fnms{P Sreenivasa} \snm{Kumar}}

%\author[B]{\fnms{Third} \snm{Author}}
\runningauthor{F. Author et al.}
\address{Department of Computer Science and Engineering, Indian Institute of Technology Madras, Chennai, India\\
E-mail: \{vinuev,psk\}@cse.iitm.ac.in }
\iffalse
\runningauthor{F. Author et al.}
\address[A]{Journal Production Department, IOS Press, Nieuwe Hemweg 6b, 1013 BG, Amsterdam,\\ The Netherlands\\
E-mail: first@somewhere.com}
\address[B]{Department first, then University or Company name, Insert a complete correspondence (mailing) address,
Abbreviate US states, Include country\\
E-mail: \{second,third\}@somewhere.com}
\fi
\begin{abstract}
Semantics-based knowledge representations such as ontologies are found to be very useful in automatically generating meaningful factual questions. Determining the difficulty-level of these system generated questions is helpful to effectively utilize them in various educational and professional applications. The existing approaches for finding the difficulty-level of factual questions are very simple and are limited to a few basic principles. We propose a new methodology for this problem by considering an educational theory called Item Response Theory (IRT). In the IRT, knowledge proficiency of end users (learners) are considered for assigning difficulty-levels, because of the assumptions that a given question is perceived differently by learners of various proficiencies. We have done a detailed study on the features/factors of a question statement which could possibly determine its difficulty-level for three learner categories (experts, intermediates, and beginners). We formulate ontology-based metrics for the same. We then train three logistic regression models to predict the difficulty-level corresponding to the three learner categories. The output of these models is interpreted using the IRT to find the question's overall difficulty-level. The performance of the models based on cross-validation is found to be satisfactory and, the predicted difficulty-levels of questions (chosen from four domains) were found to be close to their actual difficulty-levels determined by domain experts.  Comparison with the state-of-the-art method shows an improvement of 8.5\% in correctly predicating the difficulty-levels of benchmark questions.

\end{abstract}

\begin{keyword}
Difficulty-level estimation\sep Item response theory\sep Question generation
\end{keyword}

\end{frontmatter}

\renewcommand\vec[1]{\overrightarrow{#1}}
\newcommand\cev[1]{\overleftarrow{#1}}
\newtheorem{example}{Example}
\def\ss{\nobreak\hspace{.16667em plus .08333em}}

\section{Introduction}
A considerable amount of effort has been invested into the creation of a semantics-based knowledge representations such as ontologies where information is formalized into machine-interpretable formats. Among these are SNOMED CT\footnote{http://www.snomed.org/}, BioPortal\footnote{http://bioportal.bioontology.org/},  Disease ontology\footnote{http://www.berkeleybop.org/ontologies/doid.owl}, to name a few, which capture domain-specific knowledge. Given these knowledge repositories, the opportunity for creating automated  systems which utilize the underlying knowledge is enormous. Making use of the semantics of the information, such systems could perform various intelligently challenging operations. For example, a challenging task which often required in an e-Learning system  is to generate questions about a given topic which match the end users' (learners') educational need and their proficiency level.

%9446376971

The problem of generating question items from ontologies has recently gained much attention in the computer science community~\cite{asma,ontogen,mining,owled,zito,swrl}. This is mainly due to the utility of the generated questions in various educational and professional activities, such as   learner assessments in e-Learning systems, quality control in human computational tasks and, fraud detection in crowd-sourcing platforms~\cite{Seyler3}, to name a few.  

Traditionally, question generation (QG) approaches have largely focused on retrieving questions from raw text, databases and other non-semantics based data sources. However, since these sources do not capture the semantics of the domain of discourse,  the generated questions cannot be machine-processed, making them less employable in many of the real-world applications.  For example, questions that are generated from raw text are suitable only for language learning tasks~\cite{tahaniphd}. Using semantics-based knowledge sources in QG has various advantages, such as (1) in ontologies, we model   the semantic relationships between domain entities, which help in generating meaningful and machine-processable questions  (2) ontologies  enable standard reasoning and querying services over the knowledge, providing a framework for generating  questions more easily. 

Many efforts in the ontology-based QG are accompanied by methods for automating the task of difficulty-level estimation.  In the E-ATG system~\cite{EV2016SWJ}, a state-of-the-art QG system,  we have proposed an interesting method for predicting difficulty-level of the system generated factual questions. To recall, in that method, we assign a relatively high difficulty score to a question, if the concepts and roles in the question form a rare combination/pattern. For example, considering movie domain,  if a question contains the roles: \emph{is based on} and \emph{won oscar}, which rarely appear together,  the question is likely to be more difficult than those questions which are formed using a common role combination, say, \emph{is directed by} and \emph{is produced by}. Even though this method can correctly predict the difficulty-levels to a large extent, there are cases where this method fails. This is because there are other factors which influence the difficulty-level of a question.  

An early effort to identify factors that could potentially  predict the difficulty-level was  by Seyler et. al~\cite{Seyler1,Seyler2}. They have introduced a method to classify a question as \emph{easy} or \emph{hard} by finding  the features of the similar question entities in the Linked Open Data (LOD). Feature values for the classification task are obtained  based on the connectivity of the question entities in the LOD. We observed that, rather than mapping to LOD -- which is not always possible in the case of highly specific domains/domain-entities --  incorporating domain knowledge in the form of terminological axioms and following an educational theory called Item Response Theory (IRT), the prediction can be made more accurate. 

The contributions of this paper can be listed as follows.
\begin{itemize}
\item We reformulate some of the existing factors/features and propose new factors which influence the difficulty-level of a question, by taking into account the learners' knowledge level (or learners' category).  
\item We introduce ontology-based metrics for finding the feature values.
\item With the help of standard feature selection methods in machine learning and by using a test dataset, we study the influence of these factors in predicting hardness of a question for three standard learner categories. 
\item We then propose three learner-specific regression models trained only with the respective influential features and,  the output of the models is  interpreted using the IRT  to find the overall difficulty-level of a question.  
\end{itemize}

This paper is organized as follows. Section 2 contains the preliminaries  required for understanding the paper. Section 3 discusses the outline of the proposed method. In Section~\ref{rw}, we give an account of the related works. Section~\ref{factor} proposes the set of features of a question which determines its difficulty-level. In Section~\ref{ml}, we explain the machine learning methods that we have adopted to develop the  Difficulty-level Model (DLM).  Further, we discuss the performance of DLM in Section~\ref{regx}. A comparison with the state-of-the-art method is given in Section~\ref{compx}. Conclusions and future line of research are detailed at the end.

%=====================PRELIMINARIES====================
%======================================================

\section{Preliminaries}
We assume the reader to be familiar with Description Logics\cite{Baader} (DLs). DLs are decidable fragments of first-order logic with the following building blocks: unary predicates (called \emph{concepts}), binary predicates (called \emph{roles}), instances of concepts (called \emph{individuals}) and  values in role assertions (called \emph{literals}). A DL ontology is thought of as  a body of knowledge describing some domain using a finite set of DL axioms. The concept assertions and role assertions form the assertion component (or ABox) of the ontology. The concept inclusion, concept equality, role hierarchy etc. (the type of axioms depend on the expressivity of the DL) form the terminological component (or TBox) of the ontology.

\subsection{Question generation using patterns\label{patternx}}

 For a detailed study of difficulty-level estimation, we use the \emph{pattern-based} method, employed in the E-ATG system, for generating factual questions from the ABox of the given ontologies. 
  
In the pattern-based question generation,  a question can be considered as a set of \emph{conditions} that asks for a solution which is explicitly present in the ontology. The set of conditions  is formed using different combinations of concepts and roles assertions associated with an individual in the  ontology. Example-\ref{eg1} is on such  question,  framed from the following assertions that are associated with the (\emph{key}) individual \texttt{birdman}. 

\texttt{Movie(birdman)}

 \texttt{isDirectedBy(birdman,alejandro)}

\texttt{hasReleaseDate(birdman,"Aug 27 2014")}

\begin{example}\normalfont \label{eg1}\vskip -0.5em
Name the \emph{Movie} that \emph{is directed by Alejandro} and \emph{has release date {Aug 27, 2014}}.%\\
\end{example}

For generating a question of the above type, we may need to use a (generic) SPARQL query  template as shown below. The resultant tuples are then associated with a question pattern (E.g., Name the [?C], that is [?R1] [?o1] and [?R2] [?o2]. (key: ?s)) to frame the questions.
{\small
\begin{verbatim}
SELECT ?s ?C ? R1 ?o1 ?R2 ?o2   WHERE
{    
    ?s a ?C .  ?s ?R1 ?o1 .  ?s ?R2 ?o2 .
    ?R1 a owl:ObjectProperty . 
    ?R2 a owl:DatatypeProperty .   
}
\end{verbatim}
}

In~\cite{EV2016SWJ}, the authors  have studied all the possible generic question patterns that are useful in generating common factual questions. 
They have also proposed methods for selecting \emph{domain-relevant} resultant tuple55s/questions  for conducting domain related assessments.  A resultant tuple of the above query (for example, \texttt{?s = birdman, ?C = Movie, ?R1 = isDirectedBy, ?o1 = alejandro, ?R2 = hasReleaseDate, ?o2 = "Aug 27 2014"}) can be represented in the form of a set of triples (\texttt{\{(birdman, a, Movie), (birdman, isDirectedBy, alejandro), (birdman, hasReleaseDate, "Aug 27 2014")\}}). These triples, without the key, give rise to concept expressions that represent the conditions in the question. For example, the concept expression of \enquote{\texttt{(\_\_\_, a, Movie)}} is the concept \texttt{Movie} itself. Similarly, the concept expression of \enquote{\texttt{(\_\_\_, isDirectedBy, alejandro)}} is \texttt{$\exists$isdirectedBy.\{alejandro\}}. The  conditions for the  question given in Example-1 are:
\noindent\begin{itemize}\small\setlength{\itemindent}{-.6cm}
\item[]{\bf Conditions: }\texttt{Movie, $\exists$isdirectedBy.\{alejandro\},\\ $~~~~~~~~~~\exists$hasReleaseDate.\{"Aug 27 2014"\}}
\end{itemize}
It should be noted that, \texttt{\small$\exists${directedBy.}\{alejandro\}} does not imply  that the movie is directed \emph{only} by Alejandro, but it is mandatory that he should be a director of the movie.

For the ease of understanding, all examples presented in this paper are from the Movie domain.

\subsection{Item Response Theory\label{irt}}
Item Response Theory (IRT)~\cite{sage}  models relationship between the ability or trait of a person   and his responses to the \emph{items}  in an experiment. The term \emph{item} denotes  an entry, statement or a question used in the experiment. The item response can be \emph{dichotomous} (yes or no; correct or incorrect; true or false) or \emph{polytomous} (more than two options such as rating of a product). The quality measured by the item may be knowledge proficiency, aptitude, belief or even attitude. 
This theory was first proposed in the field of psychometrics, later, the theory was employed widely in educational research to calibrate and evaluate questions items in the world-wide examinations such as  the Scholastic Aptitude Test (SAT) and Graduate Record Examination (GRE)~\cite{IRT2}.  

In our experiments, we use the simplest IRT model often called \emph{Rasch model} or the \emph{one-parameter logistic model} (1PL)~\cite{IRT}. According to this model, a learner's response to a question item is determined by her knowledge proficiency level (a.k.a. \emph{trait level}) and the difficulty of the item. 1PL is expressed in terms of the probability that a learner with a particular trait level will correctly answer a question  that has a particular difficulty-level.~\cite{sage} represents this model as:
\begin{equation}\label{hdns}
\begin{split}
P(R_{li} = 1|\theta_{l},\alpha_{i})=\frac{e^{(\theta_{l} - \alpha_{i})}}{1 + e^{(\theta_{l} - \alpha_{i})}}
%\nonumber
\end{split}
\end{equation}

In the equation, $R_{li}$ refers to the response ($R$) made by the learner $l$ for the question item $i$ (where $R_{li}=1$ refers to a correct response), $\theta_{l}$ denotes the trait level of the learner $l$, $\alpha_{i}$ represents the difficulty score of item $i$. $\theta_{l}$ and  $\alpha_{i}$ values are normalized to be in the range [-1.5 to 1.5].  $P(R_{li} = 1 | \theta_{l},\alpha_{i})$ denotes the conditional probability that a learner $l$ will respond to item $i$ correctly. For example, the probability that a  below-average trait level (say, $\theta_{l} = -1.4$) learner  will correctly answer a question that has a relatively high hardness (say, $\alpha = 1.3$) is:
\begin{equation}\label{egdd}
\begin{split}
P=\frac{e^{(-1.4 - 1.3)}}{1 + e^{(-1.4 - 1.3)}}=\frac{e^{(-2.7)}}{1 + e^{(-2.7)}}=0.063
\nonumber
\end{split}
\end{equation}
%In our work, we intend to find the $\alpha_{i}$ of the factual questions which are meant for learners, whose trait levels are known to be either high, medium or low. The trait levels of the learners are found by gathering (and normalizing) their grades or marks obtained for a standard test of subject matter conducted in their enrolled institutions (see Section~\ref{empi} for details). The corresponding $P$ values are obtained by finding the ratio of the number of learners (in the trait level under consideration) who have correctly answered the item,  to the total number of learners at that trait level. On getting the values for $\theta_{l}$ and $P$, the value for $\alpha_{i}$ was calculated using the   Equation-\ref{vv}.
In the paper, we intend to find the $\alpha_{i}$ of the factual questions which are meant for learners, whose trait levels are known to be either high, medium or low.  We find the trait levels of the learners  by gathering (and normalizing) their grades or marks obtained for a standard test of subject matter conducted in their enrolled institutions. The corresponding $P$ values are obtained by finding the ratio of the number of learners (in the trait level under consideration) who have correctly answered the item,  to the total number of learners at that trait level. On getting the values for $\theta_{l}$ and $P$, the value for $\alpha_{i}$ was calculated using the   Equation-\ref{vv}.
\begin{equation}\label{vv}
\begin{split}
\alpha_{i}=\theta_{l}- log_{e}(\frac{P}{1-P})
%\nonumber
\end{split}
\end{equation}

In the equation, $\alpha_{i} = \theta_{l}$, when $P$ is 0.50.
That is, a question's difficulty is defined as the trait level required for a learner to have 50 percent probability of answering the question item correctly. Therefore, for a trait level of $\theta_{l}=1.5$, if $\alpha_{i}\approx 1.5$, we can consider that the question as having a high difficulty-level. Similarly, for a trait level of $\theta_{l}=0$, if $\alpha_{i}\approx 0$, the question has a medium difficulty-level. In the same sense, for a trait level of $\theta_{l}=-1.5$, if $\alpha_{i}\approx -1.5$, then question has a low difficulty-level.

%======================BACKGROUND========================
%==========================================================

\section{Outline of the proposed method\label{bg}}

In this paper, based on the insights obtained by the study of the questions that are generated from the  ATG\cite{flairs15} and E-ATG systems, we  propose features/factors that can positively or negatively influence the difficulty-level of a question.  Albeit there are existing methods which utilize some of these factors for predicting difficulty-level, studying the psychometric aspects of these factors  by considering learners' perspective about the question, has given us further insight into  the problem. 
 
 As we saw in Section~\ref{irt}, IRT is an item oriented theory which could be used to  find the difficulty-level of a question by knowing the question's hardness (difficult or not difficult) with respect to various learner categories. Therefore, on finding the hardness of a given question  based each on learner category,  we can effectively use the IRT model for interpreting its overall difficulty-level. 

 According to IRT, a question is assigned a \emph{high} difficulty-level if it is difficult for an expert learner to answer it correctly. A question is said to be difficult for an expert if the probability of a group of expert learners answering the question correctly is $\le 0.5$. {Similarly, a question can be assigned a \emph{medium} and \emph{low} difficulty-level if the probability with which the question is answered by a group of intermediate learners  is $\le 0.5$ and a group of beginner level learners is $\le 0.5,$ respectively.}   Table~\ref{dla} shows the difficulty-level assignment of three questions: $Q_{1}, Q_{2}$ and $Q_{3},$ based on whether they are difficult (denoted as $d$) or not difficult (represented as $nd$) for three learner categories.
 
   \begin{table}[!h]
  \centering\caption{Assigning one of the three difficulty-levels: \emph{high, medium} and \emph{low},  by considering whether  the question is difficult ($d$) or not-difficult ($nd$) for three learner categories.\label{dla}}
 \begin{tabular}{ccccl}\toprule
       Qn.          & Expert & Intermed. & Beginner & Difficulty\\
                   &  &  &  & -level\\\midrule
  $Q_{1}$ & $d$ & $d$ & $d$ & $high$\\
  $Q_{2}$ & $nd$ & $d$ & $d$ & $medium$\\
  $Q_{3}$ & $nd$ & $nd$ & $d$ & $low$\\\bottomrule
 \end{tabular}
 \end{table}
  % \end{minipage}}~~
   %\begin{minipage}{5 cm}
  \begin{figure}[h!]
  \centering
  \includegraphics[width=0.42\textwidth]{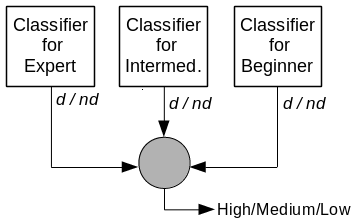}
   \caption{ Block diagram of the proposed model for predicting a question's difficulty-level\label{classifiers}}
%\end{minipage}
   \end{figure} 
 
 We consider  three standard categories of learners: \emph{beginners, intermediates}  and \emph{experts}, and model three classifiers for predicting the difficulty corresponding to the three learner categories,  as shown in Fig.~\ref{classifiers}.  Since 
  the hardness ($d/nd$)  corresponding to the three categories of learners should be predicted first from the feature values,   machine learning models/classifiers which can learn from available training data is an obvious choice.   
  We consider only those factors which are influential for a given learner category for training the models. The output of the three classifiers is matched with the content of  Table~\ref{dla} to find the question's overall difficulty-level.

%  \begin{figure}[th!]
%\scalebox{1}{
%\begin{minipage}{ 6.5 cm}

%================RELATED WORK======================
%===================================================

\section{Related Work: Difficulty-level Estimation\label{rw}}
A simple notion to find the difficulty-level of an ontology-generated multiple choice questions (MCQs) was first introduced by Cubric and Tosic\cite{Cubric2010}. Later, in~\cite{simieee}, Alsubait et al. extended the idea and proposed a similarity-based theory for controlling the difficulty of ontology-generated MCQs.  In~\cite{mining}, they have applied the theory on analogy type MCQs. In~\cite{lesson}, the authors have experimentally verified their approach in a student-course setup.  The practical solution which they have suggested to find out the difficulty-level of an MCQ is with respect to the degree of similarity of the distractors to the key. If the distractors are very similar to the key,  students may find it very difficult to answer the question, and hence it can be concluded that the MCQ is difficult.

In many a case, the question statement in an MCQ is also a deciding factor for the difficulty of an MCQ. For instance, the predicate combination or the concepts used in a question can be chosen such that they can make the MCQ difficult or easy to answer. This is the reason why  in this paper we focus on  finding difficulty-level of questions having no choices (i.e., non-MCQs). An initial investigation of this aspect was done in~\cite{EV2016SWJ}. Concurrently, there was another relevant work by Seyler et. al\cite{Seyler1,Seyler2}, focusing on QG from knowledge graphs (KGs) such as DBpedia. For judging the difficulty-level of such questions, they have designed a classifier trained on Jeopardy! data. The classifier features were based on statistics computed from the KGs (Linked Open Data) and Wikipedia. 
However, they have not considered the learner's knowledge level, as followed in the IRT, while formulating the feature metrics. This makes their measures less employable in sensitive applications such as in an e-Learning system. While considering ontology-based questions, one of the main limitation of their approach is that the feature values were determined based on the connectivity of question entities in the KG, whereas in the context of DL ontologies, the terminological axioms can be also incorporated to derive more meaningful feature metrics. In addition, the influence of the proposed factors in determining the difficulty using feature selection methods was not studied.

\section{Proposed Factors to determine Difficulty-level of Questions\label{factor} }
In this section, we look at a set of   factors which can possibly influence the difficulty-level of a question and propose ontology-based metrics to calculate them. The intuitions for choosing those factors are also detailed.

To recall, a given question can be thought of as a set of conditions. For example, consider the following questions (where the underlined portions denote the equivalent ontology concepts/roles used).
\begin{itemize}\small\setlength{\itemindent}{-.6cm}
\item[]{\bf Qn-1: }\emph{Name the \underline{Movie} that was \underline{directed by} Clint Eastwood}.  
\item[]{\bf Qn-2: }\emph{Name the \underline{Oscar movie} that was \underline{directed by} Clint Eastwood}. 
\end{itemize}

The equivalent set of conditions of the two questions can be written as:
\begin{itemize}\small\setlength{\itemindent}{-.6cm}
\item[]{\bf Conditions in Qn-1: }\texttt{Movie,}\\\texttt{$~~~~~~~~~~~~~~~~~\exists$directedBy.\{clint\_eastwood\}}
\item[]{\bf Conditions in Qn-2: }\texttt{Oscar\_movie,}\\\texttt{$~~~~~~~~~~~~~~~~~\exists${directedBy.}\{clint\_eastwood\}}
\end{itemize}

\subsection{Popularity} Popularity is considered as a factor because of the intuition that the greater the popularity of the entities that form the question,  more likely that a learner answers the question correctly.  (We observe that this notion is applicable for learners of all categories.) Therefore, the question becomes easier to answer if the popularity of the  concepts and roles that are present in the question is high.  For example, out of the following two questions, Qn-3 is likely to be easy to answer than Qn-4, since \texttt{Oscar\_movie} is a popular concept than \texttt{Thriller\_movie}.
\begin{itemize}\small\setlength{\itemindent}{-.6cm}
\item[]{\bf Qn-3: }\emph{Name an \underline{oscar movie}}.
\item[]{\bf Qn-4: }\emph{Name a \underline{thriller movie}}.
\end{itemize}

Our approach for measuring popularity is based on the observation that, (similar to what we see in Wikipedia data) if more articles talk about a certain entity, the more important, or popular, this entity is. In Wikipedia, when an article mentions a related entity, it is usually denoted by a link to the corresponding Wikipedia page. These links form a graph which is exploited for measuring the importance of an entity within Wikipedia. Keeping this in mind, we can define the popularity  of an entity (individual) in an ontology as the number of object properties which are linked to it from other individuals. For obtaining a measure in the interval [0,1], we divide the number of in-links by the total amount of individuals in the ontology.

To find  the popularity of a concept $C$ in  ontology $\mathcal{O},$  we find the mean of the popularities of all the individuals which satisfy $C$ in $\mathcal{O}$. If the condition in a question is a role restriction,  then the {concept expression}  of it will be considered, and popularity is calculated. The overall popularity of the question is determined by taking the mean of  the popularities of all the concepts and role restrictions present in it.

\subsection{Selectivity\label{sel}}
Selectivity of the conditions in a question helps in measuring the quality of the hints that are present in it~\cite{Seyler1}. Given a condition, selectivity refers to the number of individuals that satisfy it.  When the selectivity is high, a question tends to be easy to answer. For example, among the following questions, clearly, Qn-5 is easier to answer than Qn-6. This is because finding an actor who has acted in at least a movie is easy to answer than finding an actor who has acted in a particular movie; finding the latter requires more specific knowledge.
\begin{itemize}\vskip -.6 em\small\setlength{\itemindent}{-.6cm}
\item[]{\bf Qn-5: }\emph{Name an {actor} who acted in a movie}.  
\item[]{\bf Qn-6: }\emph{Name  an actor who acted in {Argo}}. 
\end{itemize}%\vskip -.5 em
To formalize such a notion, we can look at the \emph{answer space} corresponding to each of the conditions in the questions. Answer space simply denotes the \emph{count of individuals} satisfying a given condition. We will represent answer space of a condition $c$ as $ASpace(c).$

The conditions in the above questions are:
\begin{itemize}\vskip -2 em\small\setlength{\itemindent}{-.6cm}
\item[]{\bf Conditions in Qn-5: }\texttt{Actor, $\exists$actedIn.Movie}
\item[]{\bf Conditions in Qn-6: }\texttt{Actor, $\exists${actedIn.}\{argo\}}
\end{itemize}%\vskip -.6 em

Since {\small$ASpace($\texttt{$\exists${actedIn.}\{argo\}})} is very much lesser than % $<<\\ 
{\small $ASpace($\texttt{$\exists$actedIn.Movie}),} we can say that Qn-6 is difficult to answer than Qn-5. (Actors who acted only in dramas are not possible answers to Qn-5.)

As a question can have more than one conditions present in it, answer spaces of all the condition have to be taken into account while calculating the overall difficulty score of the question. It is debatable that including a specific condition in the question can always make the question difficult to answer -- sometimes a specific condition can give a better hint to a (proficient) learner.

For example, the following question is more difficult to answer than Qn-5 and Qn-6 for a non-expert, since {\small$ASpace$(\texttt{American\_actor}) $<<$ $ASpace$(\texttt{Actor})}. \vskip -2 em
\begin{itemize}\small\setlength{\itemindent}{-.6cm}
\item[]{\bf Qn-7: }\emph{Name an American  actor who acted in {Argo}}. 
\end{itemize}%\vskip -.6 em
However, for an expert,  given that the actor is an American is an additional hint, making the question sometimes easier than Qn-5 and 6. Therefore, we can roughly assume the relation between difficulty-level and answer space as follows, where $D_{expert}$ and $D_{beginner}$ correspond to the difficulty-level for an expert learner and difficulty-level for a beginner respectively. We will closely look at these relations in the following subsections. 
\vskip -2 em
%\begin{minipage}{5 cm}
\begin{equation}\label{DRequ1}
D_{expert}\propto {ASpace}
\nonumber
\end{equation}\vskip -2 em
%\end{minipage}
%\begin{minipage}{7 cm}
\begin{equation}\label{DRequ2}
D_{beginner}\propto \frac{1}{ASpace}
\nonumber
\end{equation}
%\end{minipage}

When a question contains multiple conditions, we do an aggregation of their normalized (or relative) answer spaces (denoted as $RASpace$) to find the overall answer space (addressed as $ASpaceOverall$) of the question. We find the $RASpace$ of a concept by dividing the count of individuals satisfying the concept by the total count of individuals in the apex concept (Thing class) of the ontology. For instance, $RASpace$(\texttt{\{argo\}}) = $ASpace$(\texttt{\{argo\}})/$ASpace$(\texttt{owl:Thing}). Similarly, if the condition is a role related restriction, corresponding domain concept of the role will be used to find the relative answer space. For  {\small\texttt{\small$\exists${actedIn.}\{argo\}}, $RASpace$} is calculated as:  {\small$ASpace($\texttt{$\exists${actedIn.}\{argo\}})/$ASpace($Domain(\texttt{actedIn}))}. The overall answer space can be found by taking the average of all the relative answer spaces of the conditions in the question, where $C_{S}=\{t_{1}, t_{2},..., t_{n}\}$ is the set of conditions in the question $S$, and $|C_{S}|=n.$\vskip -.7 em
\begin{equation}\label{Dxyz}\small
ASpaceOverall(C_{S}) =  \frac{\sum_{i=1}^{n} RASpace(t_{i})}{n}
\end{equation}\vskip -.3 em

In the following paragraphs, we discuss how the selectivity feature would affect the difficulty-level of an item. We discuss the cases of expert, intermediate and beginner learners separately.  In the process, we define two selectivity based features and specify how to compute them using the knowledge base and the domain ontology. 

\paragraph{\bf Expert learner} 

\begin{figure}%{r}{6.5cm}
%\vskip -4em
 \centering
  \includegraphics[width=.55\textwidth]{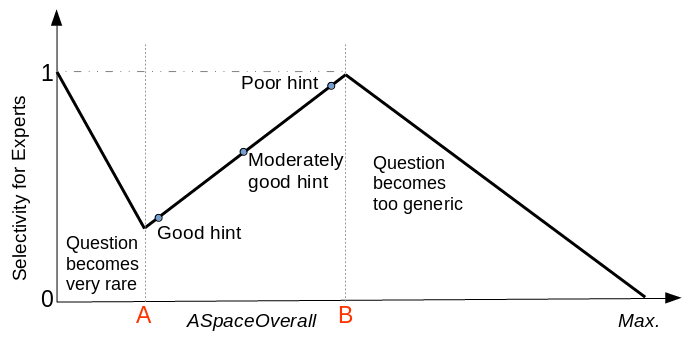}
   \caption{  Relation between   selectivity and answer space  for experts\label{test1}}
%\vskip -3 em
\end{figure} 

An expert learner is assumed to have a well developed structured knowledge about the domain of discourse. She is supposed to clearly distinguish the  terminologies of the domain and is capable of  doing reasoning over
 them. Therefore, in general, selectivity can be assumed to be directly proportional to the difficulty-level; that is, when the \emph{ASpaceOverall} increases, the underlying hints becomes poor and the question is likely to become difficult for her. 
However, intuitively,  below and beyond particular \emph{ASpaceOverall} values, a question's difficulty does not necessarily follow this proportionality.  As pointed out in~\cite{EV2016SWJ,flairs15} when a question pattern becomes rare, it becomes difficult to answer the question correctly. Therefore, in Fig.~\ref{test1}, towards the left of the point A, the question tends to become difficult, since the answer space becomes too small. Similarly, towards the right of the point B, the question tends to become more generic and its difficulty diminishes.  To accurately predict whether a question is difficult or not, it is necessary to statistically determine the positions of the points A and B. 
 Based on the initial analysis of  the empirical data obtained from~\cite{EV2016SWJ},  we processed with an assumption that the question tends to become too generic when the $ASpaceOverall$ $\ge 50\%$ of the total number of individuals in the ontology. Similarly, the question starts to become difficult when the $ASpaceOverall$ $\le 10\%$ of the total number of individuals. The selectivity corresponding to an expert is expressed as \emph{Selectivity$_{Ex}$}. Knowing the overall answer space of a question, selectivity is computed directly from the graph in Fig.~\ref{test1} -- in the graph, Max, A(10\%) and B(50\%) are known points.

   \paragraph{\bf Beginner learner\label{DMB}} A beginner is assumed to have a less developed internal knowledge structure. She can be assumed to be familiar with the generic (sometimes popular) information about the  domain and is less aware about the detailed  specifics. 
\begin{figure}%{r}{4.7cm}
%\vskip -2em
 \centering
  \includegraphics[width=.38\textwidth]{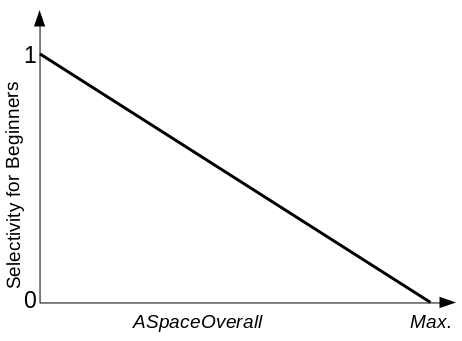}
   \caption{ Relation between  selectivity and answer space for beginners\label{test2}}
%\vskip -2.2em
\end{figure} 
    We assume that the \emph{selectivity} factor  behaves proportionally to the \emph{ASpaceOverall}, unlike what we saw in the experts' case. The intuition behind this assumption is that, when the overall answer space increases, as in the case of an expert the so-called hints in the question cannot be expected  to become poor; this is because, a person with poorly developed domain knowledge may not be able to  differentiate the quality or property of the hint, making it rather a factor for generalizing the question (thereby making the question easily answerable). Therefore, we can follow a linear proportionality relation as shown in Fig.~\ref{test2}, to find the difficulty for  a beginner, and we can denote this  new selectivity as \emph{Selectivity}$_{Bg}$.

  \paragraph{\bf Intermediate learner\label{DMI}}
An intermediate learner can be assumed to have partially both the perspective of an expect as well that of a beginner. Therefore, we can assume her selectivity value as combination  of  \emph{Selectivity}$_{Ex}$ and \emph{Selectivity}$_{Bg}$ -- considering them as two factors.

\subsection{Coherence}
In the current context, coherence captures the semantic relatedness of  entities (individuals and concepts) in a question. It can be best compared to measuring the co-occurrences of individuals and concepts in the text. While considering coherence as a factor, we  assume that higher the coherence between individuals/concepts in a question, lower is its difficulty-level and vice versa, because  intuitively, the facts about highly coherent entities are likely to be recalled easier than the facts about less coherent entities. It is observed that this notion is applicable for learners of all categories.

\begin{itemize}\small\setlength{\itemindent}{-.6cm}
\item[]{\bf Qn-8: }\emph{Name the {hollywood-movie} {starring} Anil Kapoor and Tom Cruise.} 
\item[]{\bf Qn-9: }\emph{Name the  hollywood-movie starring Tom Cruise and Tim Robbins}.  
\end{itemize}
Considering  the above two questions, coherence between  the concept \texttt{HollywoodMovie} and the individuals: \texttt{anil\_kapoor, tom\_cruise}, is lesser (since there is only one movie they both have acted together) than the coherence between \texttt{HollywoodMovie, tom\_cruise} and \texttt{tim\_robbins}, making the former question difficult to answer than the latter.

    Given an ontology, we measure the coherence between two of its individuals   
    as the sum of the ratio between the size of the set of entities that point to both individuals and  the size of the union of the sets of entities that point to either one of the individuals, and the ratio  between the size of the set of entities that are pointed by both individuals and  the size of the union of the sets of entities that are pointed by either one of the individuals. Formally, the coherence  between two individuals $p$ and $q$ can be represented as in Eq.~\ref{dd1}, where $I_{i}$ is the set of entities from which the individual $i$ is having   incoming relations and $O_{i}$  is the set of entities to which $i$ is having  outgoing relations.
\begin{equation}\label{dd1}
Coherence(p,q) = \frac{|I_{p}\cap I_{q}|}{|I_{p}\cup I_{q}|}+\frac{|O_{p}\cap O_{q}|}{|O_{p}\cup O_{q}|}
\end{equation}    
 Each portion of the measure is known as the Jaccard similarity coefficient, which is a statistical method to compare the similarity of sets.

\subsection{Specificity}
Specificity refers to how specific a question is. For example, among the following questions, Qn-2 is more specific question than Qn-10 and requires more knowledge proficiency to answer it correctly. We consider Qn-2 as more difficult to answer than Qn-10.
\begin{itemize}\small\setlength{\itemindent}{-.6cm}
\item[]{\bf Qn-2: }\emph{Name an \underline{Oscar movie} that was \underline{directed by} Clint Eastwood}.  
\item[]{\bf Qn-10: }\emph{Name the \underline{movie} that \underline{is related to} Clint Eastwood}.  
\end{itemize}

 For a  learner, the difficulty-level depends on how detailed the question is. Intuitively, if a question contains domain specific conditions, the probability of a learner for correctly answering the question will reduce. (This notion is observed to be applicable for all categories of learners.) To capture this notion, we utilize the concept   and role hierarchies in the domain ontology. We relate the depths of the concepts and roles that are used in the question to the concept and role hierarchies of the ontology, to determine the question difficulty.  
To achieve this, we introduce \emph{depthRatio} for each predicate $p$ in an ontology. \emph{depthRatio} is defined as:
\begin{equation}\label{DRequ}\small
\emph{depthRatio}_\mathcal{O}(p)= \frac{
\splitfrac{
\text{~~~~~~Depth (or length) of $p$}
}{\text{from the  root of the hierarchy~~}}
}{
\splitfrac{\text{~~~Maximum  length of }}
{ \text{the path containing $p$~~~~~~}}
}
%\nonumber
\end{equation}

For a question $S$, generated from an ontology $\mathcal{O}$, with $x$ as key and $P$ as the set of concepts/roles in $S$, let $\mathcal{C}$ denote the set of concepts satisfied by $x$, and let $\mathcal{R}$ represents the set of roles such that either $x$ is present at their domain (subject) or range (object) position \big(i.e., $R \in \mathcal{R} \implies \mathcal{O}\models R(x,i) \lor R(i,x)$, where $i$ is an arbitrary instance in $\mathcal{O}$\big). For each $p\in P$, we find the largest subset in $\mathcal{C}$  (if $p$ is a concept) or we find the largest subset in $\mathcal{R}$ (if $p$ is a role), such that the elements in the subset can be related using the relation $\sqsubseteq$, and $p$ is an element in that subset. The cardinality of such a subset forms the denominator of Eq.~\ref{DRequ}, and the numerator is the position of the predicate $p$ from the right (right represents the top concept or top role) when the elements in the subset are arranged using the relation $\sqsubseteq$.

%A question can have more than one predicate present in it. To define the overall depthRatio of the question (called the \emph{specificity}) we find the predicates which are associated with the key having the maximum depthRatio and, multiply the depthRatio with the average of all the depthRatios. We assume that the presence of predicate with a high depthRatio (associated with the reference individual) can make the question more specific. The specificity of a question $S$  can be computed as the product of the average depthRatio with the maximum of all the depthRatios.
A stem can have more than one predicate present in it. In that case, we assume that the   predicate with a highest depthRatio (associated with the reference individual) could potentially make the stem more specific.  Therefore, we define the overall depthRatio of a stem (called the \emph{specificity}) as 
 the product of the average depthRatio with the maximum  of all the depthRatios.

\section{  Difficulty-level Modeling of Questions\label{ml}}
In the previous section, we have proposed a set of features  which possibly influence  the difficulty-level of a question. In this section, we do a feature selection study using three widely used  filter models to find out the amount of influence of the proposed factors in  predicting  question difficulty. We then  train three logistic regression models ($RM_{e}, RM_{i}, RM_{b}$)  for each  learner category (experts, intermediates and beginners, respectively) using the selected prominent features. Their predictions for a given question are taken to find the overall difficulty-level.  Ten-fold cross validation is used to find the performance of the three models. %We do not report a comparison with the model proposed in~\cite{Seyler1,Seyler2} because their difficulty-level model is not a domain ontology-based model and prediction is possible only if the question components can be mapped to Linked Open Data entities. In addition, they could predict the question difficulty either as \emph{easy} or \emph{hard}, whereas our model classifies the question into three standard difficulty-levels: \emph{high, medium} and \emph{low}. 

%\noindent\emph
\paragraph{\bf Training data}
The training data consisted of a set of 520  questions that were generated from four  ontologies (DSA, MAHA, GEO and PD ontologies -- see our project
 website\footnote{Project website: https://sites.google.com/site/ontoassess/} for details) available online. These questions were classified as \emph{difficult} or \emph{not-difficult}  for each of the three learner categories  (we denote the training data for experts, intermediate and beginners respectively as $TD_{e}, TD_{i}$ and $TD_{b}$). The classification  is done by either of the two ways, 
 \begin{figure}%{r}{5cm}\vskip -2.5 em
%{\small\vskip -1 em
\begin{verbatim}
Item identifier: dsa_1
Popularity: 0.231
Selectivity_Ex: 0.320
Selectivity_Bg: 0.113
Coherence: 0.520
Specificity: 0.440
Difficulty: d
\end{verbatim}%\vskip -1.9 em
%}
\caption{An instance of the training data\label{data}}%\vskip -1.5 em
\end{figure}
(1) in a classroom setting  by using IRT or (2) with the help of  subject matter experts. In the former case, we find the probability by which a particular question is answered correctly by a learner of  specific knowledge proficiency level and assign  it as difficulty
 ($d$) or not ($nd$). In the latter case, more than 5 domain experts were asked to do the 
 ratings and their majority ratings were considered for assigning  $d$ or $nd$.  All the question that  were used for training had been previously used as benchmark sets in~\cite{Vinu2015,EV2016SWJ,flairs15}. In the training data, the question identifiers are accompanied by  five feature values tabulated from the respective ontologies along with their difficulty assignment. The feature values are normalized to  values between 0 and 1. An instance of the training data is given in Fig.~\ref{data}.

\paragraph{\bf Feature Selection}
In order to find out the amount of influence of each of the proposed factors, we did an attribute evaluation study using  three popular feature selection approaches:  {Information Gain\cite{IGFR} (IG), ReliefF\cite{RFilter} (RF) and Correlation-based\cite{CBFS} (CB)} methods. These feature selection approaches select a subset of features that minimize redundancy and maximize relevance to the target such as the class labels in classification. The ranking scores/weights obtained  for the features are given in Table~\ref{rank}.

\begin{table*}
\centering\caption{\hbox{\parbox[l][.8cm][c]{16cm}{Ranking score of features  for the three training sets using three popular filter models. (IG, RF and CB, denote the three filter models: Information Gain, ReliefF and Correlation-based, respectively.)}}}\label{rank}

%\caption{Ranking score of features  for the three training sets using three popular filter models. (IG, RF and CB, denote the three filter models: Information Gain, ReliefF and Correlation-based, respectively.)\label{rank}}%\vskip -1em
%\begin{tabular}{l@{~~}ccc@{$~~$} ccc@{$~~$}ccc}\toprule
\begin{tabularx}{\textwidth}{@{}lXXX@{~~~}XXX@{~~~}XXX@{}}\toprule
&\multicolumn{3}{c}{\bf IG} &\multicolumn{3}{c}{\bf RF} &\multicolumn{3}{c}{\bf CB}\\
&$TD_{e}$&$TD_{i}$&$TD_{b}$&$TD_{e}$&$TD_{i}$&$TD_{b}$&$TD_{e}$&$TD_{i}$&$TD_{b}$\\\midrule
Popularity                 &
  0.7613  &
  0.6344 & 
  0.7925 &  
  0.831  &  
  0.378 & 
  0.178 &  
  0.766  & 
  0.621 & 
   0.562\\

Selectivity$_{Ex}$  &  
0.8802  &  
{0.6913} & 
\cellcolor[gray]{0.8}{0.0963} &  
0.738  &  
\cellcolor{red!25}{0.451} &
\cellcolor[gray]{0.8}{0.120} & 
0.699  &  
\cellcolor{red!25}{0.258} & 
\cellcolor[gray]{0.8}{0.058}\\

Selectivity$_{Bg}$  & 
\cellcolor{blue!25}{0.0012} &  
0.6553 & 
0.9996 &  
\cellcolor{blue!25}{0.008} & 
{0.668} & 
 0.177 &  
 \cellcolor{blue!25}{0.114} &  
 0.442 & 
 0.249\\

Coherence   &
 0.5638 & 
 0.3251 & 
 0.8112 &  
 0.761  & 
 0.487 &
 0.211 &  
 0.731  & 
 0.538 & 
 0.315\\

Specificity         &
 0.6577  & 
 0.5436 & 
 0.5751 &  
 0.651  & 
 0.521 & 
 0.459 &  
 0.657  & 
 0.761 & 
 0.602\\\bottomrule
\end{tabularx}%
%\vskip -1.8em
\end{table*}

In Table~\ref{rank}, we can see that, the least prominent feature for finding the difficulty for experts %are: Coherence, Popularity, Selectivity$_{Ex}$ and Specificity; 
is the Selectivity$_{Bg}$, since all the three filter models ranked it as the least influential one -- see the fields shaded in blue in the three $TD_{e}$ columns. % is found to have the least influence. 
In the case of predicting difficulty for intermediates, the ranking scores of Selectivity$_{Ex}$ is  less  than that of Selectivity$_{Bg}$ when the models used are RF and CB -- see the fields shaded in red.  When it comes to beginner learners, the factor Selectivity$_{Ex}$ is found to have the least influence -- see the fields shaded in gray. While developing the IDM, we have ignored the least influential features for training the regression models.

\paragraph{\bf Observations} Consistent to what we have postulated in Section~\ref{sel}, Selectivity$_{Ex}$ is found to be a more influential factor than  Selectivity$_{Bg}$, for deciding the difficulty of a question for an expert learner. Similarly, for a beginner, Selectivity$_{Bg}$ is found to be more influential than Selectivity$_{Ex}$. 
\subsection{Performance of regression models\label{regx}}
 The performances  of three learner-specific regression models: $RM_{e}, RM_{i}, RM_{b}$,  considering all the 5 features are 76.73\%, 78.6\% and 84.23\% respectively. These percentage values indicate the ratio of the number of instances classified correctly to the total number of instances given for classification, under 10-fold cross-validation setting. After removing the least influential features, the performance of the classifiers became 76.9\%, 79.8\%, and 85\% respectively. The difference in the performance before and after feature selection is roughly the same because the model can theoretically assign minimum or zero weight to non-influential features. However, we did the feature selection and ranking to evaluate our hypothesis about what features are influential in which case.
 
 When the overall system was run on the available dataset of factual questions from different domains (520 questions), it is observed that DLM correctly classifies about 77\% of them. These questions  are relevant to the domain selected using the heuristics given in~\cite{EV2016SWJ}.
 
\subsection{Non-classifiable Questions}
 Following from what we have seen in Section \ref{bg}, the DLM could not assign a
  difficulty-level to a given question if the outcomes of the three regression models 
              \begin{figure}%{r}{7cm}
%\vskip -1.5em
 \centering
  \includegraphics[width=.5\textwidth]{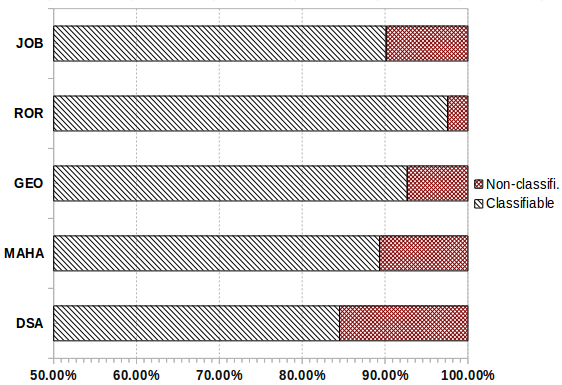}
   %\vskip -.5em
 \caption{Classifiable Vs Non-classifiable questions\label{PNP}}
%\vskip -1.9 em
\end{figure}   
  do  not   agree with the three possible assignments (see Fig. \ref{dla}). We call such questions as  \emph{non-classifiable} ones and the others as \emph{classifiable} questions. We investigated the  percentage of such non-classifiable    cases by analyzing questions generated from five ontologies available online (available in our project website).  We used questions that were generated in~\cite{flairs15} for our study, 
 the  statistics of the non-classifiable cases are given in Fig.~\ref{PNP} (Note that the X-axis begins with 50\%). On an average, 10\% of all the questions generated from an ontology are found to be non-classifiable. 
 
 An  analysis of the non-classifiable questions taken from the DSA  ontology shows that incompleteness of the ontology and the way the domain is modeled 
 as an ontology are the main reasons for this discrepancy. For example, \emph{Name a doubly linked list}  is a question which is assigned to be difficult for an expert (since the Doubly Linked List concept has only one individual and   the incompleteness of data makes it a less popular concept), and not difficult for a beginner (the depthRatio is high for the  concept \texttt{Doubly\_Linked\_List} because, it forms a specific concept in the ontology. However, this issue would not have appeared,  if the  concept \texttt{Doubly\_Linked\_List} had been modeled as an individual.
 
 \section{Comparison with existing method\label{compx}}
 In this section, we compare the predictions of difficult-levels by the proposed model and the method given in~\cite{EV2016SWJ}. We call the latter as \emph{E-ATG method}. 
 We do not report a comparison with the model proposed in~\cite{Seyler1,Seyler2} because their difficulty-level model is not a domain ontology-based model and prediction is possible only if the question components can be mapped to Linked Open Data entities. In addition, they could predict the question difficulty either as \emph{easy} or \emph{hard}, whereas our model classifies the question into three standard difficulty-levels: \emph{high, medium} and \emph{low}. 
 
 In~\cite{EV2016SWJ}, effectiveness of E-ATG method is established by comparing the predicated difficulty-levels with their actual difficulty-levels determined in a classroom setting. DSA ontology was used for the study. Twenty four representative questions  (given in Appendix~A), selected from 128213 generated questions, were utilized for the comparison.  (More details about the selection process can be found at~\cite{EV2016SWJ}).  %The questions from among the 24 questions whose (actual) difficulty-level can be determined using the item evaluation data are used as the benchmark questions. 
 A correlation of 67\% between the predicted and actual difficulty-levels was observed\footnote{To get more accurate result, the calculations were redone with $\theta$ values  between: [-1.5, 1.5], and 1.25, 0 and -1.25 as the medians of the $\alpha$ values for experts, beginners and intermediates, respectively with $\pm$.25 standard deviation}.  We tested the approach proposed in this paper on the above set of questions.
 
 While comparing  the proposed difficulty-levels with the actual difficult-levels, we found that 21 out of 24 are matching (87.5\% correlation), and one benchmark question is identified as non-classifiable.
 
 \subsection{Discussion} 
The E-ATG method mainly considered only one feature, the \emph{triviality score} (which denotes how rare the property combination in the stem are), for doing the predication. Our results (8.5\% improvement) show that  the proposed set of new features could improve the correctness of the prediction. The current model is trained only using 520 training samples. We expect the system to perform even better after training with more data as and when they are available, and by identifying other implicit features. Due to unavailability of large training data, unsupervised feature learning methods cannot be effectively applied in this context.    
 \section{Conclusions and Future Work}
 Establishing mechanisms to control and predict the difficulty of assessment questions is clearly a big gap in existing question generation literature. Our contributions have  covered %advanced the knowledge on 
 the deeper aspects of the problem, and proposed strategies, that exploit ontologies and associated measures,  to provide a better  difficulty-level predicting model,  that can address this gap. We developed the difficulty-level model (DLM) by introducing three learner-specific logistic regression models for predicting the difficulty of a given question for three categories of learners. The output of these three models was then interpreted using the Item Response Theory to assign \emph{high, medium} or \emph{low} difficulty-level.    The overall performance of the DLM and the individual performance of the three regression models based on cross-validation   were reported and they are found to be satisfactory. Comparison with the state-of-the-art method shows an improvement of 8.5\% in correctly predicating the difficulty-levels of benchmark questions.

  The model proposed in this paper for predicting the difficulty-level of questions is limited to ABox-based factual questions. It would be interesting to extend this model to questions that are generated using the TBox-based approaches. However, the challenges to be addressed would be much more, since, in the TBox-based methods, we have to deal with many  complex restriction types (unlike in the case of ABox-based methods)  and their influence on the difficulty-level of the question framed out of them needs a detailed investigation.

 For establishing the propositions and techniques stated in this paper, we have implemented a system which demonstrates the feasibility of the methods on medium sized ontologies. %However, using these systems on large ontologies depends on the availability of such ontologies. 
 It would be interesting to investigate the working of the system on large ontologies.

 \section*{Acknowledgements}
 This project is funded by Ministry of Human Resource Development, Gov. of India. We express our fullest gratitude to the participants of our evaluation process: {Dr. S.Gnanasambadan} (Director of Plant Protection, Quarantine $\&$ Storage), Ministry of Agriculture, Gov. of India; %\emph{Dr. S.Nazreen Hassan} (Asst. Professor), \emph
 {Mr. J. Delince} and {Mr. J. M. Samraj}, %, Assistant Professor(Agricultural Extension), 
Department of Social Sciences AC $\&$ RI, Killikulam, Tamil Nadu, India; {Ms. Deepthi.S} (Deputy Manager), Vegetable and Fruit Promotion Council Keralam (VFPCK), Kerala, India; {Dr.\, K.Sreekumar} (Professor) and students,  College of Agriculture, Vellayani, Trivandrum, Kerala, India.  We also thank all the undergraduate and post-graduate students of Indian Institute of Technology, Madras, who have participated in the empirical study. %AND DEFINITELY YOUR NAME WILL BE ALSO INCLUDED. 
%We are thankful to the anonymous reviewers as well as the internal reviewers (\emph{Mr.\,S. Baskaran}, \emph{Ms. Subhashree.S}, \emph{Mr. Dileep Kumar} and \emph{Mrs. Athira}) of our paper, for their valuable suggestions. 
\bibliography{ref.bib}{}
\bibliographystyle{plain}

\section*{Appendix A}\label{app:A}
Tables~\ref{egmcqs1}-\ref{egmcqs3} contain benchmark stems that are generated from the Data Structures and Algorithms (DSA) ontology. In those tables, the stems 1 to 6, 7 to 16 and 17 to 24 correspond to high, medium and low (actual) difficulty-levels respectively. These stems %along with their choice sets (please refer to our project website for choice sets) 
were employed in the experiment mentioned in Section~\ref{compx}. Stems- 7, 8 and 20 are the uncorrelated ones. Stem-20 is identified as non-classifiable  questions by our approach. \\~\\

\begin{table}[th!]%\caption{Questions  generated from the DSA ontology that are  having \emph{high} actual difficulty-level\label{egmcqs1}}
\caption{\parbox[l][.8cm][c]{7.7cm}{Questions  generated from the DSA ontology that are  having \emph{high} actual difficulty-levels.}}\label{egmcqs1}
\begin{tabular}{@{}rp{6cm}@{}}\toprule
Item No.& Stems of MCQs\\\midrule
1&Name a polynomial time problem with application in computing canonical form of the difference between bound matrices. \\

2.&Name an NP-complete problem with application in pattern matching and is related to  frequent subtree mining problem.  \\

3.&Name an all pair shortest path algorithm that is faster than Floyd-Warshall Algorithm.  \\

4.&Name an application of an NP-complete problem that is also known as Rucksack problem.    \\

5.&Name a string matching algorithm that is faster than Robin-Karp algorithm.\\

6.&Name a polynomial time problem that is also known as maximum capacity path problem.    \\
\bottomrule
\end{tabular}
\end{table}

\begin{table}[t]\caption{Questions generated from the DSA ontology that are having \emph{medium} actual  difficulty-levels.\label{egmcqs2}}
\begin{tabular}{@{}rp{6cm}@{}}\toprule
Item No.& Stems of MCQs\\\midrule

7. &Name an NP-hard problem with application in logistics.    \\

8.&Name the one whose worst time complexity is n exp 2 and with Avg time complexity n exp 2.\\   
9.&Name the one which operates on output restricted dequeue and operates on input restricted dequeue. \\

10.&Name the operation of a queue that operates on a priority queue. \\

11.&Name a queue operation that operates on double ended queue and operates on a circular queue.\\

12.&Name the ADT that has handling process \enquote{LIFO}.\\

13.&Name an internal sorting algorithm with worse time complexity m plus n.      \\
14.&Name a minimum spanning tree algorithm with design technique greedy method. \\   
15.&Name an internal sorting algorithm with time complexity n log n.\\
16.&Name an Internal Sorting Algorithm with worse time complexity n exp 2.\\

\bottomrule
\end{tabular}
\end{table}

\begin{table}[h!]\caption{Questions generated from the DSA ontology that are having \emph{low} actual difficulty-levels.\label{egmcqs3}}
\begin{tabular}{@{}rp{6cm}@{}}\toprule
Item No.& Stems of MCQs\\\midrule

17.&Name a file operation.\\    
18.&Name a heap operation. \\
19.&Name a tree search algorithm.  \\  
20.&Name a queue with operation dequeue.   \\ 
21.&Name a stack operation.\\
22.&Name a single shortest path algorithm.\\
23.&Name a matrix multiplication algorithm. \\    
24.&Name an external sorting algorithm. \\

\bottomrule
\end{tabular}
\end{table}

\end{document}